\documentclass[10pt,twocolumn,letterpaper]{article}

\usepackage[final]{cvpr}

\usepackage{times}
\usepackage{epsfig}
\usepackage{graphicx}
\usepackage{amsmath}
\usepackage{amssymb}
\usepackage{epigraph}

\usepackage[utf8]{inputenc}
\usepackage{multirow}
\usepackage{rotating}
\usepackage{verbatim}
\usepackage{booktabs}
\usepackage{array}
\usepackage{textcomp, gensymb}
\usepackage{diagbox}
\usepackage[T1]{fontenc}    
\usepackage{url}            
\usepackage{booktabs}       
\usepackage{amsfonts}       
\usepackage{nicefrac}       
\usepackage{microtype}      
\usepackage{setspace}
\usepackage{multirow}
\usepackage{amssymb}
\usepackage{diagbox}
\usepackage{mathtools}
\usepackage{wrapfig}
\usepackage{diagbox} 
\usepackage{wrapfig,lipsum,booktabs}
\usepackage{caption}
\usepackage{appendix}

\usepackage{color}
\usepackage{xcolor}
\definecolor{citecolor}{rgb}{0.21,0.49,0.74}
\usepackage{colortbl}
\usepackage{array}

\usepackage[pagebackref=false,breaklinks=true,citecolor=citecolor,colorlinks,bookmarks=false]{hyperref}

\definecolor{tblue}{RGB}{80,80,245}
\definecolor{tred}{RGB}{250,100,100}

\newcolumntype{x}[1]{>{\centering\arraybackslash}p{#1pt}}

\newcommand{\app}{\raise.17ex\hbox{$\scriptstyle\sim$}}

\newlength\savewidth

\usepackage{amssymb}
\usepackage{pifont}

\usepackage[pagebackref=false,breaklinks=true,citecolor=citecolor,colorlinks,bookmarks=false]{hyperref}

\usepackage{amssymb}
\usepackage{pifont}
\usepackage{adjustbox}
\usepackage{afterpage}

\definecolor{tgreen}{RGB}{32,178,170}

\definecolor{tgray}{RGB}{169,169,169}

\definecolor{tbg}{RGB}{230,245,230}
\definecolor{tbb}{RGB}{135,206,250}
\definecolor{blue2}{RGB}{0,139,139}
\definecolor{red2}{RGB}{255,127,80}

\definecolor{lightyellow}{rgb}{1,1, 0.8}
\definecolor{yellow}{rgb}{1,0.97, 0.65}
\definecolor{orange}{rgb}{1, 0.85, 0.7}
\definecolor{tablered}{rgb}{1, 0.7, 0.7}
\definecolor{tablegreen}{rgb}{0.80, 1, 0.80}
\definecolor{tablered2}{rgb}{0.8, 0.8, 1.0}

\newcommand{\cmark}{\color{tgreen}\ding{51}}%
\newcommand{\xmark}{\color{red2}\ding{55}}%

\newcommand{\modelname}{RayZer}

\def\paperID{**}
\def\confName{ICCV}
\def\confYear{2025}

\title{
\makebox[0pt][l]{\raisebox{-0.2em}{\includegraphics[height=2em]{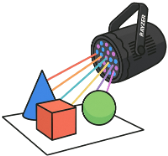}}}
\hspace{2em} \textbf{RayZer: A Self-supervised Large View Synthesis Model}
}

\vspace{-0.1in}
\author{
Hanwen Jiang$^{1}$~~~
Hao Tan$^{2}$~~~
Peng Wang$^{2}$~~~
Haian Jin$^{3}$~~~
Yue Zhao$^{1}$~~~
Sai Bi$^{2}$~~~\\
Kai Zhang$^{2}$~~~
Fujun Luan$^{2}$~~~
Kalyan Sunkavalli$^{2}$~~~
Qixing Huang$^{1}$~~~
Georgios Pavlakos$^{1}$~~~
\\
\small{
$^{1}$The University of Texas at Austin \ 
$^{2}$Adobe Research \
$^{3}$Cornell University}\\
}

\begin{document}

\twocolumn[%
\vspace*{-0.5cm}  
\maketitle
\vspace{0mm}
\begin{center}
    \captionsetup{type=figure}
    \vspace{-0.37in}
    \centering
    \includegraphics[width=\textwidth]{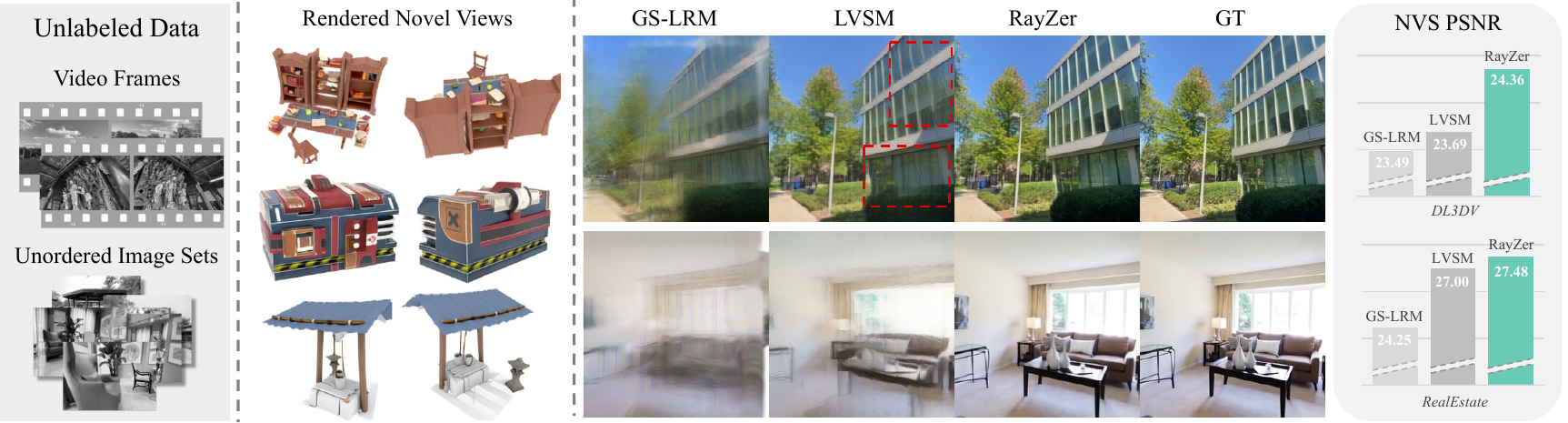}
    \vspace{-0.28in}
    \caption{We propose \textbf{\modelname{}}, a \textbf{self-supervised} multi-view 3D Vision model trained on \textit{unlabeled data} without any annotations, \eg, camera pose labels. At inference, \modelname{} supports \textit{feed-forward} novel view synthesis from \textit{unposed \& uncalibrated} images. \modelname{} achieves novel view synthesis performance comparable to that of supervised ``oracle'' methods (GS-LRM and LVSM), which require camera labels in both training and inference, and even \textbf{outperforms} them when they rely on (potentially noisy) COLMAP camera annotations. We show two examples on the right, where COLMAP camera annotations lead to consistent failures of GS-LRM and LVSM during inference.} 
    \label{fig: teaser}
\end{center}
\bigbreak
]
\afterpage{\thispagestyle{plain}}
\pagestyle{plain}


\begin{abstract}
We present \modelname{}, a self-supervised multi-view 3D Vision model trained without any 3D supervision, i.e., camera poses and scene geometry, while exhibiting emerging 3D awareness. Concretely, \modelname{} takes unposed and uncalibrated images as input, recovers camera parameters, reconstructs a scene representation, and synthesizes novel views. During training, \modelname{} relies solely on its self-predicted camera poses to render target views, eliminating the need for any ground-truth camera annotations and allowing \modelname{} to be trained with 2D image supervision.
The emerging 3D awareness of \modelname{} is attributed to two key factors. First, we design a self-supervised framework, which achieves 3D-aware auto-encoding of input images by disentangling camera and scene representations. Second, we design a transformer-based model in which the only 3D prior is the ray structure, connecting camera, pixel, and scene simultaneously. \modelname{} demonstrates comparable or even superior novel view synthesis performance than ``oracle'' methods that rely on pose annotations in both training and testing. 
Project: \href{https://hwjiang1510.github.io/RayZer/}{https://hwjiang1510.github.io/RayZer/}
   
\end{abstract}


\vspace{-0.23in}
\section{Introduction}
\vspace{-0.05in}
\label{sec: intro}

Self-supervised learning has driven the rise of foundation models, enabling training on vast amounts of unlabeled data and fueled by the scaling law~\cite{kaplan2020scaling}. This paradigm has proven highly effective for LLMs~\cite{radford2019language}, VLMs~\cite{bai2024sequential}, and visual generation~\cite{peebles2023scalable}. In contrast, 3D Vision models still rely heavily on ground-truth 3D geometry and camera pose labels~\cite{wang2024dust3r, hong2023lrm}, which are usually estimated from time-consuming optimization methods, \eg, COLMAP~\cite{schonberger2016colmap}, and are not always perfect. This reliance limits both learning scalability and effectiveness. To break free from this constraint, we move beyond the supervised paradigm and ask: \textit{how far can we push a 3D Vision model without any 3D supervision}?


In this paper, we present \textbf{\modelname{}}, a large multi-view 3D model \textbf{trained with self-supervision} and \textbf{exhibiting emerging 3D awareness}. The input of \modelname{} is \textit{unposed and uncalibrated} multi-view images, sampled from continuous video frames or unordered multi-view captures. \modelname{} first recovers the camera parameters, then reconstructs the scene representation, and finally renders novel views. The key insight of our self-supervised training is to use the camera poses \textit{predicted by \modelname{} itself} to render views that provide \textit{photometric supervision},  rather than following the standard protocol of using ground truth poses for rendering~\cite{wang2023pflrm, jiang2023leap, zhang2025flare}.
Thus, \modelname{} can be trained with \textbf{zero 3D supervision}, \ie, no 3D geometry or camera pose supervision. During inference, \modelname{} predicts camera and scene representations in a \textit{feed-forward} manner, without requiring per-scene optimization. We show inference results in Fig.~\ref{fig: teaser}.

\begin{figure}[t]
    \centering
    \includegraphics[width=\linewidth]{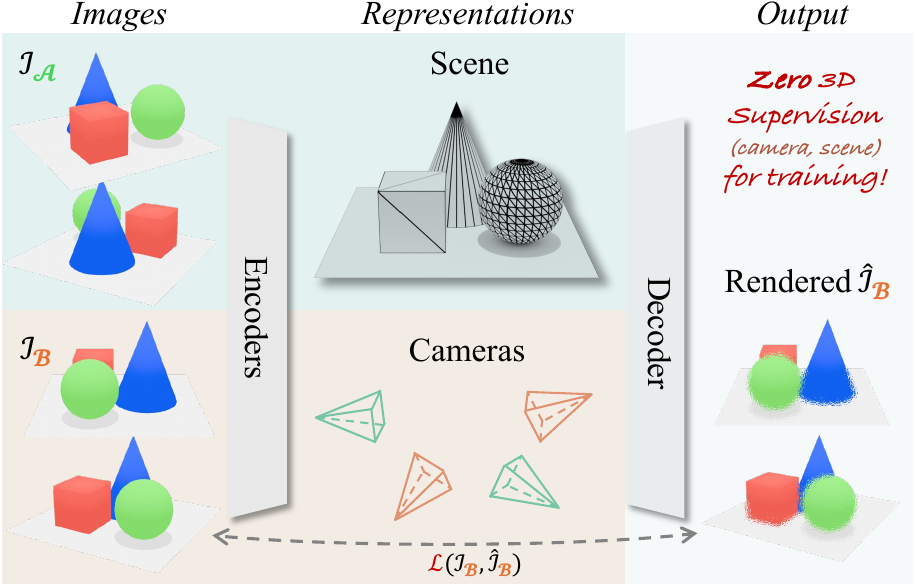}
    \vspace{-0.25in}
    \caption{\textbf{Our proposed self-supervised training framework}.
    This is an abstract design that we later operationalize with our \modelname{} model (illustrated in Fig.~\ref{fig: model} and Sec.~\ref{sec: method}). We divide the input images into two sets $\mathcal{I}_\mathcal{A}$ and $\mathcal{I}_\mathcal{B}$. We predict the scene representation from $\mathcal{I}_\mathcal{A}$, and use the predicted cameras of $\mathcal{I}_\mathcal{B}$ (shown in orange) to render the scene. We leverage photometric loss between raw input $\mathcal{I}_\mathcal{B}$ and its prediction $\hat{\mathcal{I}}_\mathcal{B}$ for training.
    } 
    \label{fig: overview}
    \vspace{-0.2in}
\end{figure}

As \modelname{} uses the camera poses predicted by itself for training, this self-supervised task can be interpreted as \textbf{3D-aware image auto-encoding}~\cite{rumelhart1985learning, zhou2017unsupervised, lai2021videoae}. Initially, \modelname{} \textit{disentangles} input images into camera parameters and scene representations (reconstruction). It then \textit{re-entangles} these predicted representations back into images (rendering). To facilitate this disentanglement, we \textbf{control the information flow}. As shown in Fig.~\ref{fig: overview}, 
we divide all images into \textit{two parts}: one set predicts the scene representation (input views), while the other offers photometric self-supervision (target views). 
This is achieved by using estimated poses of the second set to render the scene representation predicted from the first set, thereby preventing trivial solutions that are not 3D-aware.


To facilitate self-supervised learning, \modelname{} is built only with \textit{transformers} -- no 3D representation, hand-crafted rendering equation, or 3D-informed architectures. This design is motivated by self-supervised large models in other modalities~\cite{brown2020language, bai2024sequential, peebles2023scalable}, enabling \modelname{} to flexibly and effectively learn domain-specific knowledge.
The only 3D prior incorporated in \modelname{} is the \textbf{ray structure}, which simultaneously models the relationship between camera, pixels (image), and scene. Concretely, \modelname{} first predicts camera poses, which are then converted into pixel-aligned Plücker ray maps~\cite{plucker1865xvii} to guide the scene reconstruction that follows. This ray-based representation serves as a strong prior for addressing the chicken-and-egg problem of structure and motion~\cite{cornell_sfm_lecture}, effectively allowing the camera and scene representations to regularize each other during training.

We evaluate \modelname{} on three datasets, including both scene-level and object-level data with different camera configurations. We observe that \modelname{} demonstrates \textbf{comparable or even better novel view synthesis performance} than ``oracle'' methods~\cite{zhang2024gslrm, jin2024lvsm} that use pose labels in both training and testing. Interestingly, we identify that potentially noisy pose annotations from COLMAP can limit the performance of ``oracle'' models. The results not only demonstrate the effectiveness of \modelname{}, but also shows the potential of 3D Vision models to break free from supervised learning.

\section{Related Work}

\vspace{-1mm}
\noindent \textbf{Large-scale 3D Vision Models}. 3D Vision models learn 3D representations and priors from data~\cite{choy20163dr2n2, qi2016volumetric, girdhar2016learning, kar2017learning, qi2018frustum, tulsiani2018factoring, tung2019learning, zhang2022structure,zhang2024relitlrmgenerativerelightableradiance}. Recently, researchers have developed large-scale models to acquire general 3D knowledge. One research direction focuses on designing improved model architectures that incorporate the inductive biases of multi-view stereo~\cite{yu2021pixelnerf, wang2021ibrnet, chen2021mvsnerf, chen2024mvsplat} and epipolar geometry~\cite{he2020epipolar, du2023learning, charatan2024pixelsplat, chen2023explicit}. Another line of work leverages full transformer models that intentionally omit architectural 3D inductive biases~\cite{sajjadi2022srt, nichol2022point, jiang2023leap}. For example, LEAP~\cite{jiang2023leap}, LRMs~\cite{hong2023lrm, wang2023pflrm, zhang2024gslrm, wei2024meshlrm, longlrm}, and DUSt3R~\cite{wang2024dust3r, leroy2024grounding, duisterhof2024mast3rsfm, yang2025fast3r, wang2025continuous} are the first works employing transformers to convert 2D images into 3D representations. 
SRT~\cite{sajjadi2022srt} and LVSM~\cite{jin2024lvsm} further replace 3D representations and physical rendering equations with latent representations and learned rendering functions, improving performance and scalability. However, they still require ground-truth camera poses for supervised training and/or accurate camera annotations during inference. To achieve scalable supervised learning, MegaSynth~\cite{jiang2024megasynth} and Stereo4D~\cite{jin2024stereo4d} leverage synthetic data and stereo videos to expand the data scale, however, curating data for different tasks can be laborious. In contrast, \modelname{} explores self-supervised training to break free from supervised learning.

\vspace{0.02in}
\noindent \textbf{Self-supervised 3D Representation Learning}. Learning 3D-aware representations from unlabeled image data is a long-standing problem in 3D Vision. 
One line of work leverages single-view images. However, they either only work for a specific category~\cite{yan2016perspective, kanazawa2018learning, nguyen2019hologan, chan2021pi, lin2020sdfsrn, mustikovela2020self} or can only recover partial observations~\cite{chan2022efficient, sargent2023vq3d, xiang20233d}. Some works explore semi-supervised learning and achieve better scalability~\cite{yang2024depth, jiang2024real3d}, but performance is still highly restricted to the model weights, which are initialized by fully supervised training~\cite{yang2025depthv2}. The most relevant work is self-supervised learning from multi-view images~\cite{tian2020contrastive, wu2023mcc, weinzaepfel2022croco}. For example, Zhou et al.~\cite{zhou2017unsupervised}, Lai et al.~\cite{lai2021videoae}, and their following works~\cite{fu2023mononerf, ze2023visual} use camera motion as 2D or 3D warping operations to regularize learning. However, this strong inductive bias limits the learning effectiveness. RUST~\cite{sajjadi2023rust} is a pioneering work in learning latent scene representations from unposed imagery. \modelname{} is different in three aspects. First, \modelname{} initially estimates camera poses and uses poses to condition the following latent reconstruction. In contrast, RUST operates in an inverse pipeline -- it first reconstructs the scene and then estimates the camera poses. Second, \modelname{} employs different explicit pose representations to improve information disentanglement and 3D awareness, enabling novel view synthesis by geometrically interpolating predicted poses. Instead, RUST uses a latent pose representation, which makes scene-pose disentanglement challenging and they are not explicitly 3D-aware.
Third, \modelname{} follows the model architecture of LVSM~\cite{jin2024lvsm} using pure self-attention in transformers, which is different from RUST that follows SRT using convolution and cross-attention.


\begin{figure*}[htp]
    \centering
    \includegraphics[width=\textwidth]{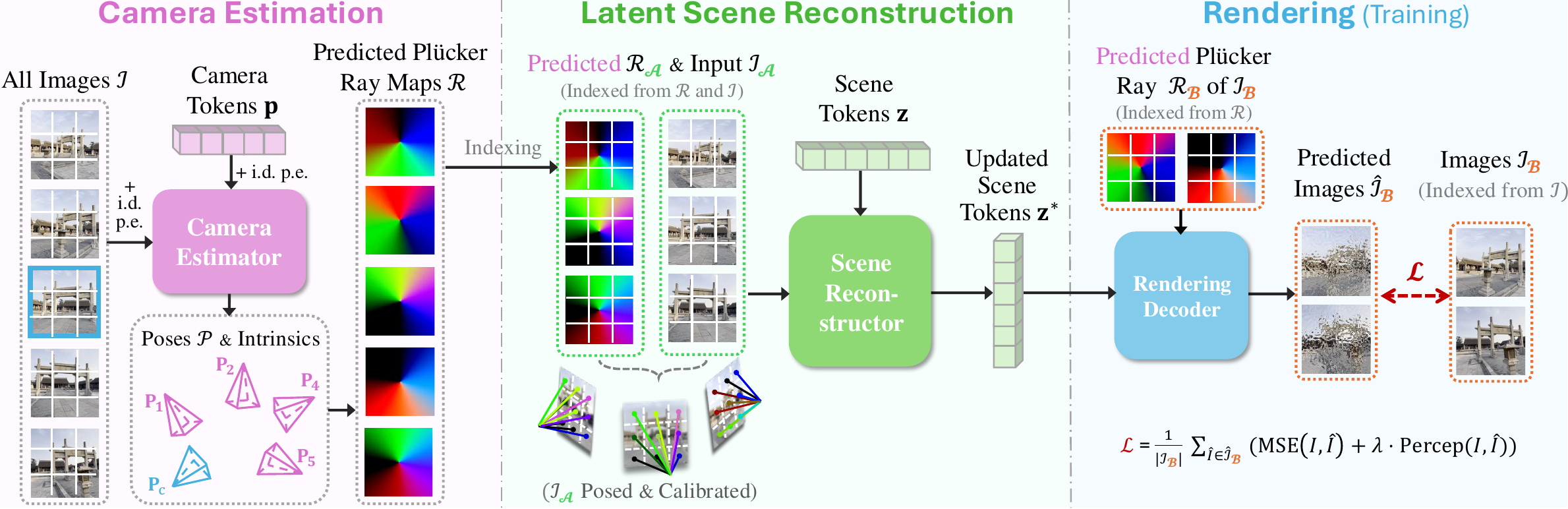}
    \vspace{-0.28in}
    \caption{\small{\textbf{\modelname{} self-supervised learning framework.} 
    } \modelname{} takes in unposed and uncalibrated multi-view images $\mathcal{I}$ and predicts per-view camera parameters and a scene representation, which supports novel view rendering. (\textbf{Left}) \modelname{} first estimates camera parameters, where one view is selected as the canonical reference view (in blue box). \modelname{} predicts the intrinsics and the relative camera poses $\mathcal{P}$ of all views. The predicted cameras are then converted into pixel-aligned Plücker ray maps $\mathcal{R}$. (\textbf{Middle}) \modelname{} uses a subset of input images, $\mathcal{I}_\mathcal{A}$, as well as their previously predicted camera Plücker ray maps, $\mathcal{R}_\mathcal{A}$, to predict a latent scene representation. Here, the Plücker ray maps, $\mathcal{R}_\mathcal{A}$, serve as an effective condition for scene reconstruction. (\textbf{Right}) \modelname{} can render a target image given the scene representation $\mathbf{z}^{*}$ and a target camera. During training, we use $\mathcal{R}_\mathcal{B}$, which is the previously predicted cameras Plücker ray maps of $\mathcal{I}_\mathcal{B}$, to render $\hat{\mathcal{I}}_\mathcal{B}$. This allows training \modelname{} end-to-end with self-supervised photometric losses between inputs $\mathcal{I}_\mathcal{B}$ and their renderings $\hat{\mathcal{I}}_\mathcal{B}$.
    } 
    \label{fig: model}
    \vspace{-0.15in}
\end{figure*}

\noindent \textbf{Optimization-based Unsupervised SfM, SLAM, and NVS}. Although these methods are not directly comparable to \modelname{}, we discuss them due to the similar input-output formulations. In detail, these methods optimize target predictions on a per-scene basis~\cite{smith1986representation, schonberger2016colmap}, while \modelname{} is a feed-forward parametrized model, learning priors by training on large data. The traditional SfM, SLAM, and NVS methods are unsupervised~\cite{schonberger2016colmap, heigl1999plenoptic}.
Although generally performing well, they are restricted by the complicated hand-crafted workflow, leading to requirements of dense-view inputs~\cite{zhang2022relpose}, slow speed~\cite{schonberger2016colmap}, and sensitivity to hyper-parameters~\cite{truong2023sparf}. Recent optimization-based NeRF and 3DGS works can also perform NVS from unposed images~\cite{lin2021barf, bian2023nope, fu2024colmapfree, smith2024flowmap}. However, they do not have learnable model parameters to encode priors, thus requiring off-the-shelf models trained with 3D supervision as regularization or providing initialization. 
\section{Preliminaries}
\label{sec: preliminary}
We introduce two important building blocks of \modelname{}, \ie, the latent set scene representation and how to render it.

\noindent \textbf{Latent set scene representation}. Compressing data into tokens in latent space is a common practice in text, image, video, etc. Recently, this representation has also been extended to 3D research~\cite{sajjadi2022srt, zhang20233dshape2vecset, sajjadi2023rust, jin2024lvsm}. In contrast to classical explicit (\eg, meshes and point clouds), implicit (\eg, NeRF~\cite{mildenhall2021nerf} and SDF~\cite{park2019deepsdf}), and hybrid (\eg, triplane~\cite{chan2022efficient} and 3DGS~\cite{kerbl20233dgs}) representations that are 3D-aware, the latent set representation is \textit{not explicitly 3D-aware}. It serves as a \textit{compression} of scene information, where the 3D-awareness properties are \textit{fully learned}. The latent set scene representation can be denoted as $\mathbf{z} \in \mathbb{R}^{n\times d}$, where $n$ is the number of tokens in the set and $d$ is the latent feature dimension. 

\noindent \textbf{Rendering latent set scene representation} requires a network, say $R^\theta$, as introduced by SRT~\cite{sajjadi2022srt} and LVSM~\cite{jin2024lvsm}. We formulate it as $v =R^\theta(\mathbf{z}, r)$, where $r$ is a ray and $v$ is the rendered property, \eg, RGB values, of the corresponding pixel\footnote{For improved efficiency and performance, LVSM groups rays from the same image patch and decodes them jointly.}. 
This formulation is the same as traditional Graphics rendering techniques~\cite{appel1968some, kajiya1986rendering}, as $v = R(\textsc{scene}, \textsc{ray})$, where $R$ is a pre-defined and handcrafted rendering equation, \eg, alpha-blending ray marching in NeRF. Differently, our ``rendering equation'' is a learned model parameterized with weights $\theta$, and our scene representation is a latent token set as discussed previously. We omit the model parametrization, \eg, weight $\theta$, in the following description for clarity.

\section{\modelname{}}
\label{sec: method}

In this section, we first introduce \modelname{}'s self-supervised learning framework (Sec.~\ref{sec: method_task}). Then, we present the details of the \modelname{} model architecture (Sec.~\ref{sec: method_model}). 

\subsection{\modelname{}'s Self-supervised Learning}
\label{sec: method_task}

We first formulate the input and output of \modelname{}. We then introduce the self-supervised learning framework.

We focus on the standard setting of modeling static scenes~\cite{schonberger2016colmap}. The \textbf{input} of \modelname{} is a set of \textit{unposed and uncalibrated} multi-view images $\mathcal{I} = \{ I_i \in \mathbb{R}^{H\times W\times 3} \vert i=1,...,K \}$, which can come from unlabeled video frames or image sets. The \textbf{output} is a parametrization of the inputs, \ie, camera intrinsics, per-view camera poses, and scene representation, enabling novel view synthesis. To predict these representations, we build the \modelname{} model and train it with \textit{self-supervised learning} -- no 3D supervision, \ie, 3D geometry, and camera pose annotations during training. 

To train \modelname{} with self-supervision, we control the data information flow. We split the input images $\mathcal{I}$ into \textbf{two non-overlapping subsets} $\mathcal{I}_\mathcal{A}$ and $\mathcal{I}_\mathcal{B}$, where $\mathcal{I}_\mathcal{A} \cup \mathcal{I}_\mathcal{B} = \mathcal{I}$ and  $\mathcal{I}_\mathcal{A} \cap \mathcal{I}_\mathcal{B} = \emptyset$. \modelname{} uses $\mathcal{I}_\mathcal{A}$ to predict the \textit{scene representation}, and use \textit{$\mathcal{I}_\mathcal{B}$ for providing supervision}. Thus, \modelname{} renders images that correspond to $\mathcal{I}_\mathcal{B}$, denoted as $\hat{\mathcal{I}}_\mathcal{B}$, and we apply photometric losses:
\begin{align}
\label{eq: loss}
    \mathcal{L} = \frac{1}{K_\mathcal{B}} \sum_{\hat{I} \in \hat{\mathcal{I}}_\mathcal{B}} (\texttt{MSE}(I, \hat{I}) + \lambda \cdot \texttt{Percep}(I, \hat{I})),
\end{align}
where $K_\mathcal{B} = \vert \mathcal{I}_\mathcal{B} \vert$ is the size (number of images) of $\mathcal{I}_\mathcal{B}$ , $I \in \mathcal{I}_\mathcal{B}$ is the image that corresponds to a predicted image $\hat{I}$, and $\lambda$ is the weight for perceptual loss~\cite{johnson2016perceptual, li2020crowdsampling}. The two sets are randomly sampled during training. 

\subsection{\modelname{} Model}
\label{sec: method_model}

\noindent \textbf{Overview}. As introduced in Sec.~\ref{sec: method_task}, \modelname{} recovers both camera parameters and the scene representation from unposed, uncalibrated input images. A key design element of \modelname{} is its \textbf{cascaded} prediction of camera and scene representations. This is motivated by the fact that even noisy cameras can be a strong condition for better scene reconstruction~\cite{schonberger2016colmap, jiang2024forge, zhang2025flare}, which is analogous to traditional structure-from-motion methods~\cite{schonberger2016colmap} and is in contrast with recent reconstruction-first methods~\cite{wang2024dust3r, wang2023pflrm, sajjadi2023rust}. This design can provide mutual regularization of predicting pose and scene during training, facilitating self-supervised learning.

\modelname{} builds a pure transformer-based model, benefiting from its scalability and flexibility.
As shown in Fig.~\ref{fig: model}, \modelname{} first tokenizes input images and uses a transformer-based encoder to predict camera parameters of \textit{all views}. In this step, the cameras are represented by their intrinsics and $\texttt{SE(3)}$ camera poses.
This \textbf{low-dimensional}, \textbf{geometrically well-defined} parametrization helps disentangle image information from the camera representation.

\modelname{} then transforms the $\texttt{SE(3)}$ camera poses and intrinsics into Plücker ray maps~\cite{plucker1865xvii}, representing the predicted cameras as pixel-aligned rays. This ray-based representation captures both the 2D ray-pixel alignment and the 3D ray geometry, providing \textbf{fine-grained}, ray-level details that encapsulate the physical properties of the camera model.
The ray maps serve as a condition for improving the reconstruction stage that follows.

From the image and predicted Plücker rays of $\mathcal{I}_\mathcal{A}$, \modelname{} uses another transformer-based encoder to predict the latent set scene representation (introduced in Sec.~\ref{sec: preliminary} and detailed later).
Then, \modelname{} uses the previously estimated cameras of $\mathcal{I}_\mathcal{B}$ to predict $\hat{\mathcal{I}}_\mathcal{B}$, providing photometric self-supervision (Eq.~\ref{eq: loss}). We now formally introduce the \modelname{} model.

\vspace{0.05in}
\noindent \textbf{Image Tokenization}. For all $K$ input images $\mathcal{I} = \{ I_i \in \mathbb{R}^{H\times W\times 3} \vert i=1,...,K \}$, we patchify them into non-overlapping patches following ViT~\cite{dosovitskiy2020vit}. Each patch is in $\mathbb{R}^{s\times s\times 3}$, where $s$ is the patch size. We use a linear layer to encode each patch into a token in $\mathbb{R}^d$, leading to a patch-aligned token map $f_i \in \mathbb{R}^{h\times w\times d}$ for each image, where $h=H/s$, $w=W/s$, and $d$ is the latent dimension.

We then add positional embeddings (p.e.) to the  tokens, enabling the following model to be aware of the spatial location and the corresponding image index of each token. Specifically, we combine the sinusoidal spatial p.e.~\cite{dosovitskiy2020vit} and the sinusoidal image index p.e.~\cite{bertasius2021space} using a linear layer; note that the image index p.e. is shared among all tokens from the same image. When training on continuous video frames, these image index embeddings also encode sequential priors, which benefits pose estimation.
Finally, we reshape the token maps of all images into a set, denoted as $\mathbf{f} \in \mathbb{R}^{Khw\times d}$ (recall that the transformer is invariant to the permutation of tokens). For brevity, we will use this notation for latent token sets throughout the rest of the paper.

\vspace{0.05in}
\noindent \textbf{Camera Estimator}. The camera estimator $\mathcal{E}_{cam}$ predicts camera parameters, \ie, camera poses and intrinsics, for all input images. We use a learnable camera token in $\mathbb{R}^{1\times d}$ as the initial feature for this prediction for all views. We repeat the token $K$ times and add them with image index p.e. such that they correspond to the $K$ images. We denote this camera feature initialization as $\mathbf{p} \in \mathbb{R}^{K\times d}$. We then use the camera estimator composed of full self-attention transformer layers to update the camera tokens, as:
\begin{align}
    \{\mathbf{f}^{*}, \mathbf{p}^{*}\} = \mathcal{E}_{cam}(\{\mathbf{f}, \mathbf{p} \}),
\end{align}
where $\{\cdot, \cdot \}$ denotes concatenation along the token dimension (the union set of two token sets), and $\mathbf{f}^{*}$ and $ \mathbf{p}^{*}$ are the updated tokens. We note that $\mathbf{f}^{*}$ is not used for the following computation -- it is only used as context to update $\mathbf{p}$ in the transformer layers. For clarity, we formulate the transformer layers as follows:
\begin{align}
    \label{math: union_set}
    & \mathbf{y}^0 = \{\mathbf{f}, \mathbf{p} \}, \\
    & \mathbf{y}^{l} = \text{TransformerLayer}^l(\mathbf{y}^{l-1}), \ \text{\footnotesize $l=1,...,l_T$}\\
    & \{\mathbf{f}^*, \mathbf{p}^*\} = \text{split}(\mathbf{y}^{l_T}),
\end{align} 
where $l_T$ is the number of layers, and the $\text{split}$ operation recovers the two token sets, inverting Eq.~\ref{math: union_set}. This notation remains consistent throughout the rest of the paper.

We then predict the camera parameters for each image independently. For camera pose prediction, we follow prior works of using relative camera poses to resolve ambiguity~\cite{zhang2022relpose, jiang2024forge}. We select one view as the canonical reference (\eg, with identity rotation and zero translation), while for every non-canonical view, we predict its relative pose with respect to the canonical view. We parametrize the $\texttt{SO(3)}$ rotation using a continuous 6D representation~\cite{zhou2019continuity}, and we predict the relative pose with a two-layer MLP as follows:
\begin{align}
    p_i = \text{MLP}_\mathit{pose}([\mathbf{p}^{*}_i, \mathbf{p}^{*}_c]),
\end{align}
where $[\cdot, \cdot]$ denotes concatenation along the feature dimension, $\mathbf{p}^{*}_i$ and $\mathbf{p}^{*}_c$ (all in $ \mathbb{R}^{d}$) are the camera tokens for image $I_i$ and the canonical view, respectively. The output $p_i \in \mathbb{R}^{9}$ represents the predicted pose parameters, which are then transformed into an $\texttt{SE(3)}$ pose $\mathbf{P}_i$ for image $I_i$.

For intrinsics prediction, following prior works~\cite{guizilini20203d, lai2021videoae}, we parameterize intrinsics using a single focal length value, under the assumptions that i) the focal lengths along the x and y axes are identical, ii) all views share the same intrinsics, and iii) the principal point is at the image center. We predict the focal length using a two-layer MLP:
\begin{align}
    \text{focal} = \text{MLP}_\mathit{focal}(\mathbf{p}^{*}_c).
\end{align}
The predicted focal length is then converted into the intrinsics matrix $\mathbf{K} \in \mathbb{R}^{3\times 3}$.


\begin{figure*}[t]
    \centering
    \includegraphics[width=\textwidth]{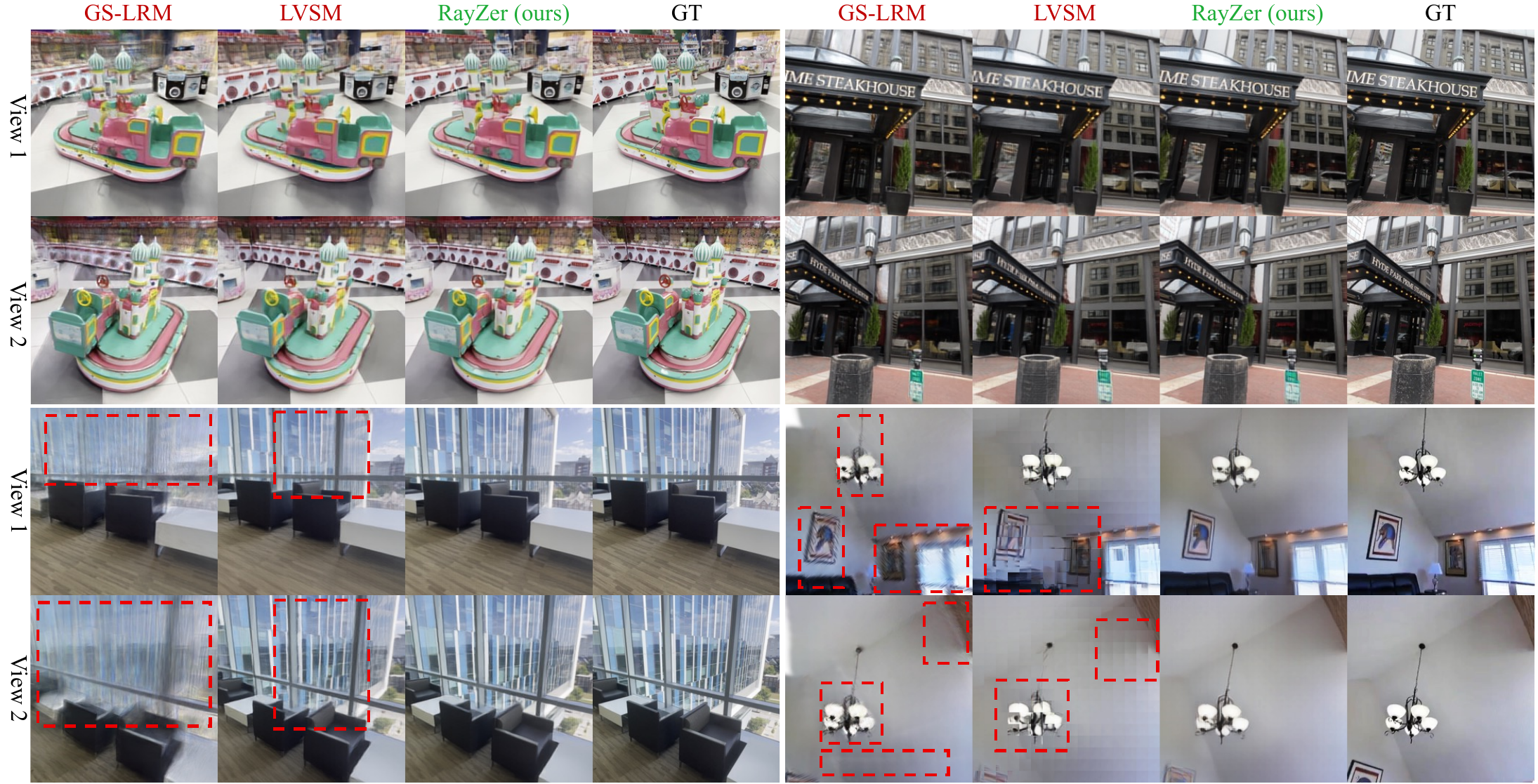}
    \vspace{-0.28in}
    \caption{\small{\textbf{Visualization results on RealEstate and DL3DV}. We compare \modelname{} with ``\textcolor[rgb]{0.75,0,0}{oracle}'' methods GS-LRM and LVSM, which use COLMAP pose annotations in both training and testing. Our \textcolor[rgb]{0.1176,0.6784,0.2196}{self-supervised} \modelname{} model does not use any pose annotations. Generally, \modelname{} performs on par with ``oracle'' methods (first row), and can outperform them on cases that COLMAP usually struggles to handle, \eg, glasses and white walls (highlighted with red boxes). The results verify our analysis on the problems of using COLMAP in Sec.~\ref{sec: main_results}. }} 
    \label{fig: vis_main}
    \vspace{-0.15in}
\end{figure*}

\vspace{0.05in}
\noindent \textbf{Scene Reconstructor}. As discussed in Sec.~\ref{sec: method_task}, we predict the scene representation from image set $\mathcal{I}_\mathcal{A}$ and additionally condition it on the previously predicted camera parameters $\mathcal{P}_\mathcal{A} = \{(\mathbf{P}_i, \mathbf{K} ) | I_i \in \mathcal{I}_\mathcal{A}\}$.
We first convert $\mathcal{P}_\mathcal{A}$ to pixel-aligned Plücker rays~\cite{plucker1865xvii} for each image, denoted as $\mathcal{R} \in \mathbb{R}^{K\times H\times W \times 6}$. Similar to image inputs, we also tokenize the Plücker rays into patch-level tokens using a linear layer, yielding $\mathbf{r} \in \mathbb{R}^{Khw\times d}$. We index the image and Plücker rays tokens corresponding to the image set $\mathcal{I}_\mathcal{A}$, denoted as $\mathbf{f}_\mathcal{A}$ and $\mathbf{r}_\mathcal{A}$ (each in $\mathbb{R}^{K_\mathcal{A}hw\times d}$, respectively). We fuse these tokens along the feature dimension with a two-layer MLP: 
\begin{align}
    \mathbf{x}_\mathcal{A} = \text{MLP}_\mathit{fuse}([\mathbf{f}_\mathcal{A}, \mathbf{r}_\mathcal{A}]),
\end{align}
where $\mathbf{x}_\mathcal{A} \in \mathbb{R}^{K_\mathcal{A}hw\times d}$ represents the fused tokens.
Importantly, we use the raw image tokens $\mathbf{f}$ rather than the pose transformer output $\mathbf{f}^*$ for this fusion. This design choice prevents leakage of information from the image set $\mathcal{I}_\mathcal{B}$, since the camera estimator transformer producing $\mathbf{f}^*$ has access to a global context that includes tokens from $\mathcal{I}_\mathcal{B}$.

We then employ a scene reconstructor $\mathcal{E}_\mathit{scene}$ consisting of full self-attention transformer layers to predict the latent scene representation. To initialize this representation, we use a set of learnable tokens $\mathbf{z} \in \mathbb{R}^{L\times d}$, where $L$ denotes the number of tokens. We formulate the process as follows:
\begin{align}
    \{ \mathbf{z}^*,  \mathbf{x}^*_\mathcal{A} \} = \mathcal{E}_\mathit{scene}(\{ \mathbf{z},  \mathbf{x}_\mathcal{A} \}).
\end{align}
The update rule is identical to the transformer layers in the camera estimator $\mathcal{E}_{cam}$. Here, $\mathbf{z}^*$ represents the final latent scene representation predicted from $\mathcal{I}_\mathcal{A}$.  Meanwhile, $\mathbf{x}_\mathcal{A}^{*}$ is discarded.


\vspace{0.05in}
\noindent \textbf{Rendering Decoder}. We first define the rendering decoder and then describe its training usage.

We use a transformer-based decoder with full self-attention for rendering, following LVSM~\cite{jin2024lvsm}. For a target image, we begin by representing it as pixel-aligned Plücker rays and tokenize these rays using a linear layer to obtain target tokens $\mathbf{r} \in \mathbb{R}^{hw \times d}$. Next, we fuse the scene information by updating the tokens with a decoder $\mathcal{D}_\mathit{render}$ comprising transformer layers:
\begin{align}
    \{ \mathbf{r}^{*}, \mathbf{z}' \} = \mathcal{D}_\mathit{render}(\{ \mathbf{r}, \mathbf{z}^{*} \}),
\end{align}
where $\mathbf{z}'$ is subsequently discarded, while the update rule of $\mathcal{D}_\mathit{render}$ is the same as previously introduced modules. Finally, we decode the RGB values at the patch level with an MLP:
\begin{align}
    \hat{I} = \text{MLP}_{rgb}(\mathbf{r}^{*}),
\end{align}
where $\hat{I} \in \mathbb{R}^{hw\times (3s^2)}$. We reshape $\hat{I}$ to recover the 2D spatial structure, yielding a final rendered image in $\mathbb{R}^{H\times W\times 3}$.

During training, we use the predicted Plücker ray maps $\mathcal{R}_\mathcal{B}$, which correspond to $\hat{\mathcal{I}}_B$, to render images of $\hat{\mathcal{I}}_B$ and then compute the self-supervised loss as defined in Eq.~\ref{eq: loss}.





\section{Experiments}
In this section, we introduce the experimental setting and present the evaluation results.
%
For the implementation,
\modelname{} employs 24 transformer layers, with 8 layers for each of the camera estimator, scene encoder, and rendering decoder. We train \modelname{} with a learning rate of $4\times 10^{-4}$ with a cosine scheduler for 50,000 iterations and a batch size of 256. The weight of perceptual loss is $\lambda=0.2$. For all experiments, we used a resolution of 256 with a patch size of 16.
More details are in the Appendix.

\subsection{Experimental Setup}
We introduce our experimental setup, including datasets, evaluation protocol and metrics, as well as baseline methods.

\vspace{1mm}
\noindent \textbf{Datasets}.
We use three datasets to evaluate \modelname{}, including two scene-level datasets, DL3DV~\cite{ling2024dl3dv} and RealEstate~\cite{zhou2018realestate}, and an object-level dataset Objaverse~\cite{deitke2023objaverse} (rendered as videos). We train and test on each dataset separately. The numbers of input views ($\mathcal{I}_\mathcal{A}$) and target views ($\mathcal{I}_\mathcal{B}$) are set to 16 and 8 for DL3DV, 5 and 5 for RealEstate, and 12 and 8 for Objaverse, respectively. We sample input images with the index ranges of 64-96, 128-192, and 50-65 on DL3DV, RealEstate, and Objaverse, respectively.
These values are chosen based on data difficulty, especially camera baseline, following prior works~\cite{charatan2024pixelsplat, zhang2024gslrm, longlrm}. We use the offical DL3DV train-test split, and split RealEstate following~\cite{charatan2024pixelsplat}.
More details can be found in the Appendix.

\noindent \textbf{Evaluation Protocol and Metrics}.
We evaluate novel view synthesis quality. Specifically, the evaluation protocol of \modelname{} is different from the ``oracle'' and supervised methods, which use ground-truth poses to render images. Instead, we use \textbf{predicted poses} to render novel views, thereby assessing the compatibility between the predicted poses and the scene representation. Since the model is trained without explicit pose annotations, the learned poses exist in a different space, and their direct correspondence to standard pose annotations is unknown. This evaluation protocol follows RUST~\cite{sajjadi2023rust}. We note that the target views are used only for pose estimation and not for scene representation prediction, ensuring that no information leakage occurs.

\noindent \textbf{Baselines}.
We compare \modelname{} with two types of methods, including 1) \textbf{``oracle'' methods}, \ie, GS-LRM~\cite{zhang2024gslrm} and LVSM~\cite{jin2024lvsm} (encoder-decoder version),
that use ground-truth camera poses during \textbf{both} training (as supervision) and inference (as pre-requisite). LVSM also uses latent set scene representation. Thus, it serves as the main comparison for the ``oracle'' methods; 2) \textbf{supervised method}, \ie, PF-LRM~\cite{wang2023pflrm}, which requires camera supervision to learn pose estimation and reconstruction; thus, it is pose-free during inference. For fair comparisons, we use 16 transformer layers in total for GS-LRM and LVSM. Thus, their number of parameters is the same as \modelname{}, except that \modelname{} has another camera estimator to handle unposed images. We use 24 transformer layers for PF-LRM.
We also consider the self-supervised method RUST~\cite{sajjadi2023rust}, but since it does not have an official public implementation, we ablate the key design differences between RUST and \modelname{} in Table~\ref{table: ablation_contribution} instead.

\begin{table}[t]
\centering
\begin{subtable}[t]{\linewidth}
    \centering
    \resizebox{1.0\linewidth}{!}{
        \begin{tabular}{l|c|c|ccc|ccc}
            \toprule
            & Training & Inference w. & \multicolumn{3}{c|}{Even Sample} & \multicolumn{3}{c}{Random Sample} \\
            & Supervision & COLMAP Cam. & PSNR$_\uparrow$ & SSIM$_\uparrow$ & LPIPS$_\downarrow$ & PSNR$_\uparrow$ & SSIM$_\uparrow$ & LPIPS$_\downarrow$  \\ 
            \midrule \midrule
            \multicolumn{9}{l}{\textbf{``Oracle'' methods} (assume inputs are posed \& use pose annotations during training)} \\
            \midrule
            GS-LRM & \cellcolor{tablered2}2D + Camera & \cellcolor{tablered2}Yes & \cellcolor{yellow}23.49 & \cellcolor{yellow}0.712 & \cellcolor{yellow}0.252 & \cellcolor{yellow}23.02 & \cellcolor{yellow}0.705 & \cellcolor{yellow}0.266\\
            LVSM & \cellcolor{tablered2}2D + Camera & \cellcolor{tablered2}Yes & \cellcolor{orange}23.69 & \cellcolor{orange}0.723 & \cellcolor{orange}0.242 & \cellcolor{orange}23.10 & \cellcolor{orange}0.703 & \cellcolor{orange}0.257\\
            \midrule \midrule
            \multicolumn{9}{l}{\textbf{Unsupervised methods} (inputs are un-posed \& no pose annotations used during training)} \\
            \midrule
            \modelname{} & \cellcolor{tablegreen}2D & \cellcolor{tablegreen}No & \cellcolor{tablered}\textbf{24.36} & \cellcolor{tablered}\textbf{0.757} & \cellcolor{tablered}\textbf{0.209} & \cellcolor{tablered}\textbf{23.72} & \cellcolor{tablered}\textbf{0.733} & \cellcolor{tablered}\textbf{0.222}\\
            \bottomrule
        \end{tabular}
    }
\end{subtable}
\vspace{-0.1in}
\captionsetup[subtable]{justification=centering}
\caption{\textbf{Evaluation results on DL3DV}. The camera annotations used by the ``oracle'' models come from \textbf{COLMAP}. The results are reported with continuous video frames (ordered) as the input. The results for the unordered image set input are in Table.~\ref{table: dl3dv_unorder}. The input and target views can be evenly or randomly sampled from video frames. We bold our result if it is better than the ``oracle'' models.} 
\vspace{-0.2in}
\label{table: dl3dv_combined}
\end{table}


\subsection{Results}
\label{sec: main_results}
\vspace{-1mm}
\noindent \textbf{Main results}. Table~\ref{table: dl3dv_combined}-\ref{table: obj} summarizes the results on the three datasets. Remarkably, without any 3D labels (e.g., camera pose annotations) during training, \modelname{} achieves performance comparable to the best “oracle” model, LVSM. In fact, \modelname{} even outperforms LVSM on DL3DV and RealEstate10k while performing slightly worse on Objaverse. We conjecture that this is because the camera poses in DL3DV and RealEstate are annotated by COLMAP, which can be imperfect and set an upper bound for “oracle” methods that are supervised by COLMAP annotations. In contrast, our self-supervised approach enables the model to learn a pose space that optimally benefits latent reconstruction and novel view synthesis. This hypothesis is further supported by the results on Objaverse -- a synthetic dataset with perfect pose annotations from the rendering tool -- where LVSM, acting as a true oracle, outperforms \modelname{}. Nonetheless, the small performance gap showcases the effectiveness of our self-supervised training. Visualizations in Fig.~\ref{fig: vis_main} further support our conjecture regarding COLMAP’s noisy poses, as both LVSM and GS-LRM consistently underperform on challenging cases that COLMAP usually fails. These results not only validate our self-supervised learning approach but also demonstrate its potential to break free from the limitations of supervised learning.

\begin{figure}[t]
    \centering
    \includegraphics[width=\linewidth]{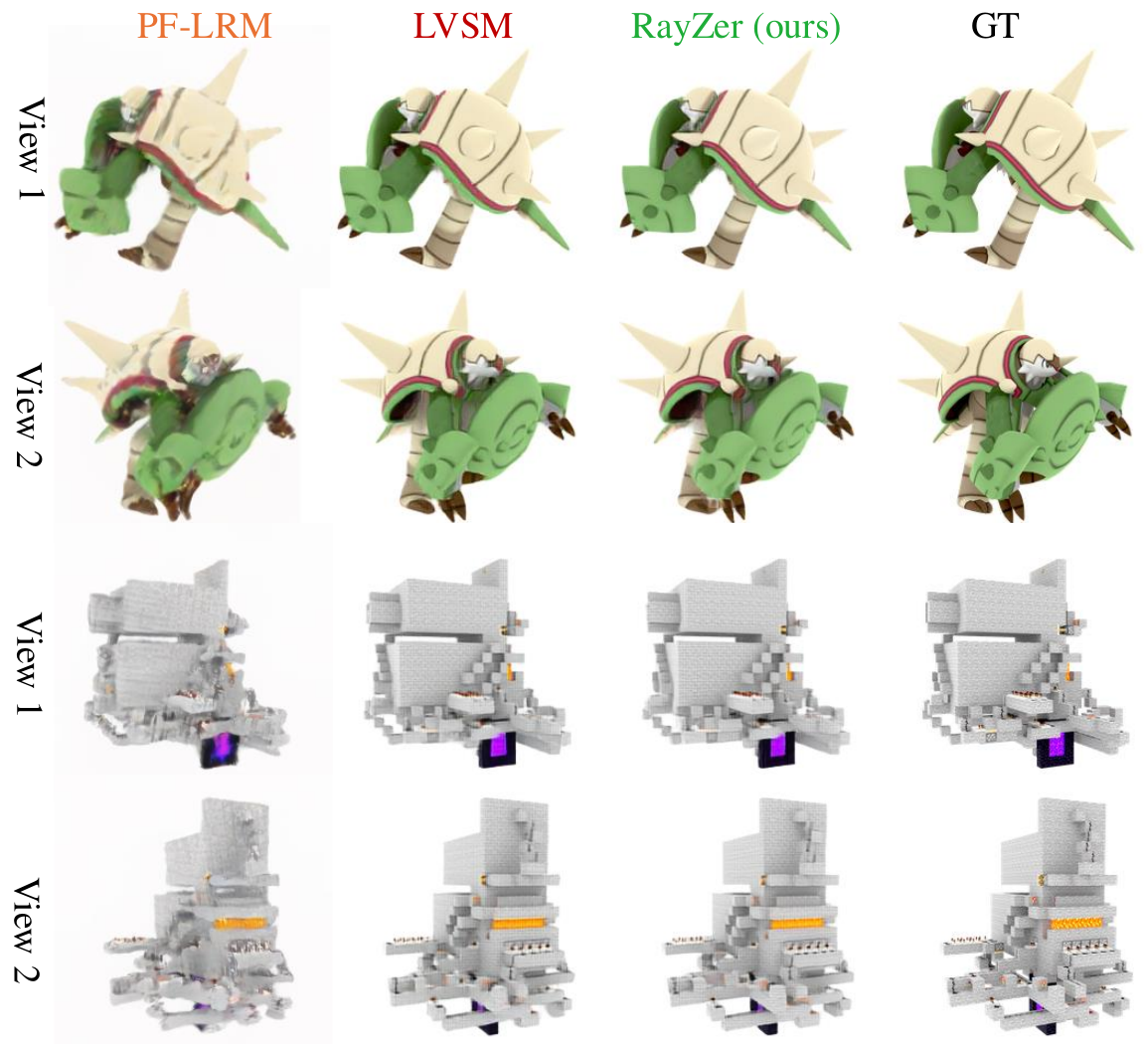}
    \vspace{-0.28in}
    \caption{\small{\textbf{Visualization results on Objaverse}. \modelname{} performs on par with LVSM and outperforms the \textcolor[rgb]{0.9137, 0.4431, 0.1961}{supervised} method PF-LRM. }} 
    \label{fig: vis_main_obj}
    \vspace{-0.06in}
\end{figure}

\begin{table}[t]
    \centering
    \resizebox{1.0\linewidth}{!}{
        \begin{tabular}{l|c|c|ccc|ccc}
            \toprule
            & Training & Inference w. & \multicolumn{3}{c|}{Even Sample} & \multicolumn{3}{c}{Random Sample} \\
            & Supervision & COLMAP Cam. & PSNR$_\uparrow$ & SSIM$_\uparrow$ & LPIPS$_\downarrow$ & PSNR$_\uparrow$ & SSIM$_\uparrow$ & LPIPS$_\downarrow$  \\ 
            \midrule \midrule
            \multicolumn{9}{l}{\textbf{``Oracle'' methods} (assume inputs are posed \& use pose annotations during training)} \\
            \midrule
            GS-LRM & \cellcolor{tablered2}2D + Camera & \cellcolor{tablered2}Yes & \cellcolor{yellow}24.25 & \cellcolor{yellow}0.770 & \cellcolor{yellow}0.227 & \cellcolor{yellow}23.21 & \cellcolor{yellow}0.748 & \cellcolor{yellow}0.251\\
            LVSM & \cellcolor{tablered2}2D + Camera & \cellcolor{tablered2}Yes & \cellcolor{orange}27.00 & \cellcolor{orange}0.851 & \cellcolor{orange}0.157 & \cellcolor{orange}25.88 & \cellcolor{orange}0.828 & \cellcolor{orange}0.175\\
            \midrule \midrule
            \multicolumn{9}{l}{\textbf{Unsupervised methods} (inputs are un-posed \& no pose annotations used during training)} \\
            \midrule
            \modelname{} & \cellcolor{tablegreen}2D & \cellcolor{tablegreen}No & \cellcolor{tablered}\textbf{27.48} & \cellcolor{tablered}\textbf{0.861} & \cellcolor{tablered}\textbf{0.146} & \cellcolor{tablered}\textbf{26.32} & \cellcolor{tablered}\textbf{0.835} & \cellcolor{tablered}\textbf{0.164} \\
            \bottomrule
        \end{tabular}
    }
    \vspace{-0.12in}
    \caption{\textbf{Evaluation results on RealEstate} with continuous video frames inputs. The camera annotations come from \textbf{COLMAP}.}
    \vspace{-0.15in}
    \label{table: realestate}
\end{table}

\noindent \textbf{Using unordered image sets for training}. \modelname{} can be trained on continuous video frames (Table~\ref{table: dl3dv_combined}-\ref{table: obj}) or unordered image sets (Table~\ref{table: dl3dv_unorder}).
Note that these two training settings are applied separately. As shown in Table~\ref{table: dl3dv_unorder}, we observe that the model trained with unordered image sets performs worse than the one trained with continuous video frames. We notice that the difference is at the pose estimation stage -- specifically, the image index positional embedding encourages local pose smoothness that benefits the learning of pose estimation on continuous frames.
This finding suggests that scaling training data using video resources, which are plentiful online, could be more advantageous than relying on unordered image sets that are often limited in scale and contain noisy content~\cite{tung2024megascenes, li2018megadepth}.

\begin{table}[t]
\centering
\resizebox{1.0\linewidth}{!}{
    \begin{tabular}{l|c|c|ccc|ccc}
        \toprule
        & Training & Inference w. & \multicolumn{3}{c|}{Even Sample} & \multicolumn{3}{c}{Random Sample} \\
        & Supervision & GT Cam. & PSNR$_\uparrow$ & SSIM$_\uparrow$ & LPIPS$_\downarrow$ & PSNR$_\uparrow$ & SSIM$_\uparrow$ & LPIPS$_\downarrow$  \\ 
        \midrule \midrule
        \multicolumn{9}{l}{\textbf{``Oracle'' methods} (assume inputs are posed \& use pose annotations during training)} \\
        \midrule
        LVSM & \cellcolor{tablered2}2D + GT Cam. & \cellcolor{tablered2}Yes & \cellcolor{tablered}32.34 & \cellcolor{tablered}0.950 & \cellcolor{tablered}0.050 & \cellcolor{tablered}32.34 & \cellcolor{tablered}0.949 & \cellcolor{tablered}0.051\\
        \midrule \midrule
        \multicolumn{9}{l}{\textbf{Supervised methods} (inputs are un-posed \& use pose annotations during training)} \\
        \midrule
        \multicolumn{1}{l|}{PF-LRM} & \cellcolor{tablered2}2D + GT Cam. & \cellcolor{tablered2}Yes (render) & \cellcolor{yellow}25.48 & \cellcolor{yellow}0.882 & \cellcolor{yellow}0.110 & \cellcolor{yellow}25.43 & \cellcolor{yellow}0.881 & \cellcolor{yellow}0.111\\
        \midrule \midrule
        \multicolumn{9}{l}{\textbf{Unsupervised methods} (inputs are un-posed \& no pose annotations used during training)} \\
        \midrule
        \modelname{} & \cellcolor{tablegreen}2D & \cellcolor{tablegreen}No & \cellcolor{orange}31.52 & \cellcolor{orange}0.945 & \cellcolor{orange}0.052 & \cellcolor{orange}31.42 & \cellcolor{orange}0.943 & \cellcolor{orange}0.053 \\
        \bottomrule
    \end{tabular}
    }
\vspace{-0.12in}
\caption{\textbf{Evaluation results on Objaverse} with continuous video frames inputs. The camera annotations are Blender ground-truth. PF-LRM uses ground-truth poses to render novel views, same with oracle methods, and we evaluate its predicted pose in Table~\ref{table: obj_interpolate_pose}. }
\vspace{-0.08in}
\label{table: obj}
\end{table}

\begin{table}[t]
\centering
\resizebox{1.0\linewidth}{!}{
\begin{tabular}{l|c|c|c|ccc|ccc}
\toprule
& Training & Inf. w. & Continuous & \multicolumn{3}{c|}{Even Sample} & \multicolumn{3}{c}{Random Sample} \\ 
& Supervision & GT Pose & Inputs & PSNR$_\uparrow$ & SSIM$_\uparrow$ & LPIPS$_\downarrow$ & PSNR$_\uparrow$ & SSIM$_\uparrow$ & LPIPS$_\downarrow$  \\ 
\midrule \midrule
(1) & 2D & No & \cmark & 24.36 & 0.757 & 0.209 & 23.72 & 0.733 & 0.222\\
(2) & 2D & No & \xmark & 20.56 & 0.576 & 0.334 & 20.02 & 0.566 & 0.356\\
\bottomrule
\end{tabular}
}
\vspace{-0.12in}
\caption{\textbf{Evaluating \modelname{} performance when using continuous or unordered images for training} on DL3DV. In evaluations, the input frames are sampled from continuous video frames. (1) keeps their temporal continuity (encoded by the image index p.e.) during training. (2) randomly shuffles the images during training.}
\vspace{-0.16in}
\label{table: dl3dv_unorder}
\end{table}

\subsection{Analysis of Camera Poses}
\begin{figure}[t]
    \centering
    \includegraphics[width=\linewidth]{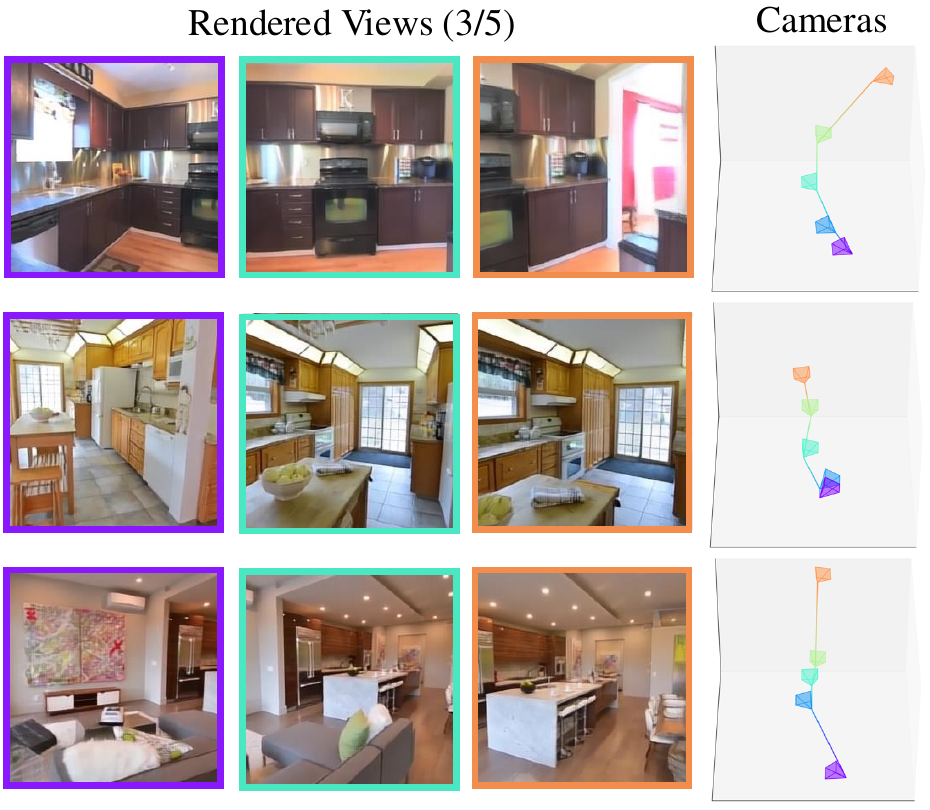}
    \vspace{-0.3in}
    \caption{\small{\textbf{Visualization of \modelname{} predicted cameras learned with self-supervision}. We visualize 3 out of 5 rendered views due to space limit, where the image index is highlighted by its color.}} 
    \label{fig: vis_cam}
    \vspace{-0.05in}
\end{figure}

\vspace{-1mm}
\noindent \textbf{\modelname{}'s learned camera pose space}. 
We visualize some camera poses predicted by \modelname{} in Fig.~\ref{fig: vis_cam}.
Although \modelname{} predicts \texttt{SE(3)} camera poses, we observe that these poses do not exactly match the real-world pose space. 
This result indicates that the \texttt{SE(3)} poses, which are later converted into Plücker ray maps, offer a degree of flexibility. Since both the rendering decoder and the scene representation operate in latent space, \modelname{} remains robust to any warping between the learned pose space and the actual real-world poses, as long as the poses are compatible with the scene representation and the decoder.

\noindent \textbf{3D Awareness of predicted camera poses}. We further investigate whether the pose space learned by \modelname{} is 3D aware. To this end, we interpolate the predicted poses of input views to synthesize more novel views, where the camera pose of a novel view is interpolated from two neighboring input views. We use ground-truth camera poses to calculate the interpolation coefficients, checking whether predicted poses follow the same geometric interpolation rules. We include the details of the interpolation method in Appendix. As shown in Table~\ref{table: obj_interpolate_pose}, \modelname{} demonstrates significantly better performance than PF-LRM and the naive baseline of copying the nearest rendered input view. These results verify that poses predicted by \modelname{} are interpolatable and 3D-aware.



\noindent \textbf{Probing the learned camera pose space.}
To probe how much actual pose information is learned by \modelname{}, we follow RUST~\cite{sajjadi2023rust} to fit a lightweight 2-layer MLP head on the pose features. 
We freeze the camera estimator’s transformers and train the MLP under camera supervision.
As shown in Table~\ref{table: tune_pose}, our probing outperforms the supervised baseline (which has the same model architecture and uses transformers trained from scratch), indicating that \modelname{}’s novel view synthesis self-supervision facilitates a better latent pose space. In contrast, supervised learning struggles due to the challenges of low-dimensional pose representation~\cite{zhang2024cameras, zhou2019continuity, levinson2020analysis, bregier2021deep, chen2022projective}.

\subsection{Ablation Study}
\vspace{-1mm}

We ablate the main design choices of \modelname{} from three aspects, including scene representation, 3D prior, and the overall model paradigm. As shown in Table~\ref{table: ablation_contribution} (1), when using the 3DGS representation rather than the latent set representation, the training does not converge. This verifies the optimization difficulty of explicit 3D representation~\cite{kerbl20233dgs, zhang2024gslrm} and demonstrates the flexibility of the latent representation with its learned rendering decoder.

Table~\ref{table: ablation_contribution} (2) and (3) ablate the prior of camera representation. Without Plücker ray maps, we observe a degraded performance in (2), showing the effectiveness of using Plücker ray maps to regularize the solution of structure-and-motion problem. Besides, we observe a slightly better performance of (3), which directly uses camera tokens $\mathbf{p}^{*}$, compared to (2). The reason is that the camera tokens $\mathbf{p}^{*} \in \mathbb{R}^{d}$ can leak target image information, while \texttt{SE(3)} poses used in (2) serve as an information bottleneck to enforce this disentanglement. Moreover, \texttt{SE(3)} poses are geometrically well-defined, allowing us to interpolate them and generate novel views along the interpolated camera trajectory, while the latent camera representation is not directly interpolable. 

Table~\ref{table: ablation_contribution} (4) ablates the overall paradigm. When the model first predicts the latent scene and then estimates poses, we observe a degraded performance. In detail, the pose estimator takes the scene representation and target image feature tokens as inputs. The result verifies our insight that pose estimation can be a strong condition for scene reconstruction, championing traditional pose-first methods in the context of self-supervised learning. Note that combining (3) and (4) will be a model that is similar to RUST conceptually.


\begin{table}[t]
\centering
\resizebox{1.0\linewidth}{!}{
    \begin{tabular}{l|c|c|ccc|ccc}
        \toprule
        & Training & Inference w. & \multicolumn{3}{c|}{Even Sample} & \multicolumn{3}{c}{Random Sample} \\
        & Supervision & GT Pose & PSNR$_\uparrow$ & SSIM$_\uparrow$ & LPIPS$_\downarrow$ & PSNR$_\uparrow$ & SSIM$_\uparrow$ & LPIPS$_\downarrow$  \\ 
        \midrule \midrule
        \multicolumn{9}{l}{\textbf{Supervised methods} (inputs are un-posed \& use pose annotations during training)} \\
        \midrule
        \multicolumn{1}{l|}{PF-LRM} & \cellcolor{tablered2}2D + GT Pose & \cellcolor{tablegreen}No & 20.63 & 0.819 & 0.160 & 21.27 & 0.827 & 0.154\\
        \midrule \midrule
        \multicolumn{9}{l}{\textbf{Unsupervised methods} (inputs are un-posed \& no pose annotations used during training)} \\
        \midrule
        \modelname{}-copy & \cellcolor{tablegreen}2D & \cellcolor{tablegreen}No & 19.56 & 0.812 & 0.159 & 20.17 & 0.820 & 0.150 \\
        \modelname{} & \cellcolor{tablegreen}2D & \cellcolor{tablegreen}No & \textbf{27.01} & \textbf{0.900} & \textbf{0.075} & \textbf{26.87} & \textbf{0.896} & \textbf{0.078}\\
        \bottomrule
    \end{tabular}
    }
\vspace{-0.12in}
\caption{\textbf{Evaluating 3D awareness of predicted camera poses} on Objaverse. Unlike Table~\ref{table: obj}, here we render novel views by interpolating predicted poses of input views, where the interpolation coefficients are calculated from GT poses. This experiment tests whether the learned $\texttt{SE(3)}$ poses are geometrically well-defined and 3D-aware. We also compare against a naive baseline ``\modelname{}-copy'' that simply copies the nearest rendered input view.}
\vspace{-0.13in}
\label{table: obj_interpolate_pose}
\end{table}

\begin{table}[t]
\centering
\resizebox{0.9\linewidth}{!}{
    \begin{tabular}{l|c|ccc|ccc}
        \toprule
        & Pose Encoder  &\multicolumn{3}{c|}{\textit{Rotation Acc.$\uparrow$ (\%)}} & \multicolumn{3}{c}{\textit{Translation Acc.$\uparrow$ (\%)}} \\
        & ($\mathcal{E}_{pose}$) & R@10$^{\circ}$ & R@20$^{\circ}$ & R@30$^{\circ}$ & t@0.1 & t@0.2 & t@0.3 \\ \midrule \midrule
        \multirow{2}{*}{DL3DV} & \cellcolor{tablered2}supervised & 39.3 & 63.0 & 77.8 & 15.7 & 33.1 & 44.4\\
         & \cellcolor{tablegreen}self-supervised & \textbf{47.6} & \textbf{72.5} & \textbf{84.0} & \textbf{20.8} & \textbf{44.0} & \textbf{60.5}\\
        \midrule
        \multirow{2}{*}{RealEstate} & \cellcolor{tablered2}supervised & 87.0 & 96.4 & 99.6 & 44.6 & 59.3 & 82.5\\
        & \cellcolor{tablegreen}self-supervised & \textbf{99.6} & \textbf{99.9} & \textbf{100} & \textbf{61.2} & \textbf{84.2} & \textbf{92.8} \\
        \midrule
        \multirow{2}{*}{Objaverse} & \cellcolor{tablered2}supervised & 19.8 & 46.7 & 66.0 & 15.1 & 37.2 & 53.8\\
        & \cellcolor{tablegreen}self-supervised & \textbf{33.6} & \textbf{69.2} & \textbf{86.8} & \textbf{20.1} & \textbf{52.7} & \textbf{75.5}\\
        \bottomrule
    \end{tabular}
    }
\vspace{-0.12in}
\caption{\textbf{Effectiveness of self-supervised pre-training for pose estimation}. We train a two-layer MLP (with supervised learning) to read out latent camera tokens $\mathbf{p}^*$ predicted by the pose encoder $\mathcal{E}_\mathit{pose}$, where the backbone is frozen. At the same time, we also compare with the baseline where both encoder $\mathcal{E}_\mathit{pose}$ and the pose prediction MLP are trained with supervised learning from scratch.}
\vspace{-0.05in}
\label{table: tune_pose}
\end{table}

\begin{table}[t]
    \centering
    \resizebox{1.0\linewidth}{!}{
    \begin{tabular}{l|l|ccc|ccc}
        \toprule
        & & \multicolumn{3}{c|}{Even Sample} & \multicolumn{3}{c}{Random Sample} \\
        & & PSNR & SSIM & LPIPS & PSNR & SSIM & LPIPS  \\ 
        \midrule \midrule
        (0) & \modelname{} & \textbf{24.36} & \textbf{0.757} & \textbf{0.209} & \textbf{23.72} & \textbf{0.733} & \textbf{0.222}\\ \midrule
        (1) & \textbf{Representation} - 3DGS + rasterization & -- & --  & \multicolumn{2}{c}{failed} & -- &  --   \\ \midrule
        (2) & \textbf{Prior} - no Plücker ray, use $\texttt{SE(3)}$ pose & 22.73 & 0.687 & 0.249 & 21.88 & 0.647 & 0.274\\
        (3) & \textbf{Prior} - no explicit pose, use latent camera & 23.13 & 0.700 & 0.251 & 22.36 & 0.668 & 0.272\\ \midrule
        (4) & \textbf{Paradigm} - scene first, not pose first &  13.31 & 0.338 & 0.732 & 13.12 & 0.337 & 0.729\\
        \bottomrule
    \end{tabular}
    }
\vspace{-0.12in}
\caption{\textbf{Ablation study of \modelname{} designs} on DL3DV with continuous inputs. (1) is a variant uses the 3D Gaussian representation rather than latent scene representation with its learned rendering decoder used by \modelname{}; (2) does not use Plücker ray maps $\mathcal{R}_\mathcal{A}$ for conditioning latent reconstruction. Instead, it encodes the $\texttt{SE(3)}$ poses $\mathbf{P}_\mathcal{A}$ and intrinsics $\mathbf{K}$ into tokens as condition; (3) directly uses the latent camera tokens $\mathbf{p}^{*}$, rather than converting it to any explicit forms of cameras, to condition the latent scene reconstruction; (4) first reconstructs latent scene and then estimates pose as Plücker ray maps, contrasting our pose-first paradigm.}
\vspace{-0.1in}
\label{table: ablation_contribution}
\end{table}

\vspace{-1mm}
\section{Conclusion}
\vspace{-1mm}
We introduce \modelname{}, a self-supervised large multi-view 3D Vision model trained with zero 3D supervision, \ie, no 3D geometry and camera annotations. \modelname{} achieves comparable or even better novel view synthesis performance than prior works that use pose labels in both training and inference, verifying the feasibility of breaking free from supervised learning.




\clearpage

{\small
\bibliographystyle{ieeenat_fullname}
\bibliography{main}
}

\newpage
\newenvironment{suppmat}{
  \begingroup
  \setcounter{section}{0}
  \renewcommand{\thesection}{\Alph{section}}
}{
  \endgroup
}

\begin{suppmat}

\section{Experimental Details}
In this section, we introduce more details of \modelname{}.

\noindent \textbf{Objaverse Data Details}. We render Objaverse as continuous videos for training and evaluation. 
The frames are rendered with corresponding cameras on a unit sphere with a constant distance to the object center.
Specifically, we render about 70 frames for azimuth 0$^{\circ}$ to 360$^{\circ}$, where the elevation is randomly sampled between -20$^{\circ}$ to 60$^{\circ}$ for each shape instance. We sample frames with the distance between the first frame and the last frame being 50 to 65, covering the camera azimuth rotation for about one cycle.

\noindent \textbf{Camera Interpolation Details}. For the experiment of interpolating predicted cameras, we use Spherical Linear Interpolation (Slerp) for interpolating the camera pose rotation. This is based on the fact that the camera of Objaverse is moving at a constant speed. Thus, Slerp ensures the correct rotation interpolation. We then find the location on the unit sphere that corresponds to this interpolated rotation angle. Thus, we ensure the interpolated cameras are still on the unit sphere, which matches the camera sampling rule for rendering. In conclusion, this interpolation assumes that 1) the camera is moving in a constant speed, and 2) the rule of sampling camera location is known. Thus, this interpolation is only applicable to the synthetic Objaverse data, and does not apply to DL3DV and RealEstate.

\noindent \textbf{More Training Details}. For all transformer layers in \modelname{}, we apply QK-Norm~\cite{henry2020qknorm} to stabilize the training. We use a latent dimension of 768 for \modelname{} and all baselines methods. \modelname{} and LVSM both use a latent set scene representation with 3072 tokens. We use mixed precision training~\cite{micikevicius2018mixed} with BF16, further accelerated by FlashAttention-V2~\cite{dao2023flashattention} of xFormers~\cite{xFormers2022} and gradient checkpointing~\cite{chen2016training}.

We train \modelname{} and all baselines with the same training protocol. We use 32 A100 GPUs with a total batch size of 256. During training, we warm up with 3000 iterations, using a linearly increased learning rate from 0 to $4e-4$. We apply a cosine learning rate decay, while the final learning rate is $1.5e-4$. We train all baselines with $50,000$ steps. We clip the gradient with norm larger than 1.0. We follow all other hyper-parameters of LVSM.

\noindent \textbf{More Model Details}. Following LVSM, we do not use bias terms in linear and normalization layers. We also apply the depth-wise initialization for transformer layers.

\noindent \textbf{Ablation details}. In Table~7 (2), we use a two-layer MLP to encode the camera pose and intrinsics back to a latent pose representation in $\mathbb{R}^{d}$. In detail, for the predicted pose of each image (in 6D representation~\cite{zhou2019continuity}), and the camera intrinsics (as the 4-dimensional focal length and principal points of x-axis and y-axis), we first concatenate them, getting a 10-dimensional pose representation. Then, we use the MLP to map it as a high-dimensional pose feature token. To predict the target views, we use a set of learnable patch-aligned spatial tokens shared across all target images as the initialization. Thus, the rendering decoder takes in the spatial tokens, the scene tokens, and the pose token. After using transformer for updating, we use the updated spatial tokens to regress the pixel values.

\section{\modelname{} Training with Continuous Inputs}
\modelname{} takes in multi-view image inputs, which can be sampled from either continuous video frames or an unordered image set.
In this section, we present two design choices to improve self-supervised learning on video frames input.

\noindent \textbf{Canonical View Selection}. Prior works~\cite{wang2023pflrm, jiang2023leap} usually select the first image in an image sequence as the canonical view. In contrast, we select the frame at the middle time-step as canonical. In this context, the pose prediction $\text{MLP}_{pose}$ initialized with a zero mean for its weights will have a small pose data variance. Otherwise, when using the first frame as canonical, the variance can be much larger. Note that this difference in pose variance can be easily handled with ground-truth camera supervision, thus, prior works choose the first image as the canonical view. However, this is more important for unsupervised methods, like \modelname{}.

\noindent \textbf{Curriculum}. We gradually increase the training difficulty by sampling video frames with an increasing distance range. With proper initialization of the model for camera pose estimation, it first learns from images with small camera baselines, benefiting the training with larger camera baselines, that follows. In detail, we use a curriculum with a frame sampling range of 48-64, 96-128, and 24-32 at the beginning of training for DL3DV, RealEstate, and Objaverse, respectively. The frame sampling range is linearly increased to 64-96, 128-192, and 48-65 at the end of training for DL3DV, RealEstate, and Objaverse, respectively. The final frame sampling range is also used for the evaluation. The sampling ranges are set based on the difficulty (camera baseline) of each dataset, following prior works~\cite{charatan2024pixelsplat, zhang2024gslrm, longlrm, jiang2024megasynth, wang2023pflrm}.

\noindent \textbf{Experiments}. We include ablations in Table~\ref{table: ablation_tricks}, where removing any of the previously discussed techniques leads to a degraded performance. This demonstrates the effectiveness of our designs of selecting canonical view and using frame sampling curriculum during training.

\begin{table}[t]
    \centering
    \resizebox{1.0\linewidth}{!}{
    \begin{tabular}{l|l|ccc|ccc}
        \toprule
        & & \multicolumn{3}{c|}{Even Sample} & \multicolumn{3}{c}{Random Sample} \\
        & & PSNR & SSIM & LPIPS & PSNR & SSIM & LPIPS  \\ 
        \midrule \midrule
        (0) & \modelname{} & \textbf{24.36} & \textbf{0.757} & \textbf{0.209} & \textbf{23.72} & \textbf{0.733} & \textbf{0.222}\\ \midrule
        (1) & first frame as canonical &  23.86 & 0.736 & 0.224 & 23.78 & 0.737 & 0.225\\ \midrule
        (2) & no curriculum & 23.87 & 0.734 & 0.226 & 23.87 & 0.735 & 0.226\\
        \bottomrule
    \end{tabular}
    }
\vspace{-0.12in}
\caption{\textbf{Ablation study of \modelname{} techniques to train on continuous video frames}. (1) is a variant choosing the first image in the sequence as the canonical view, rather than choosing the middle frame. (2) does not use the frame sampling curriculum. }
\vspace{-0.1in}
\label{table: ablation_tricks}
\end{table}

\begin{figure}[t]
    \centering
    \includegraphics[width=\linewidth]{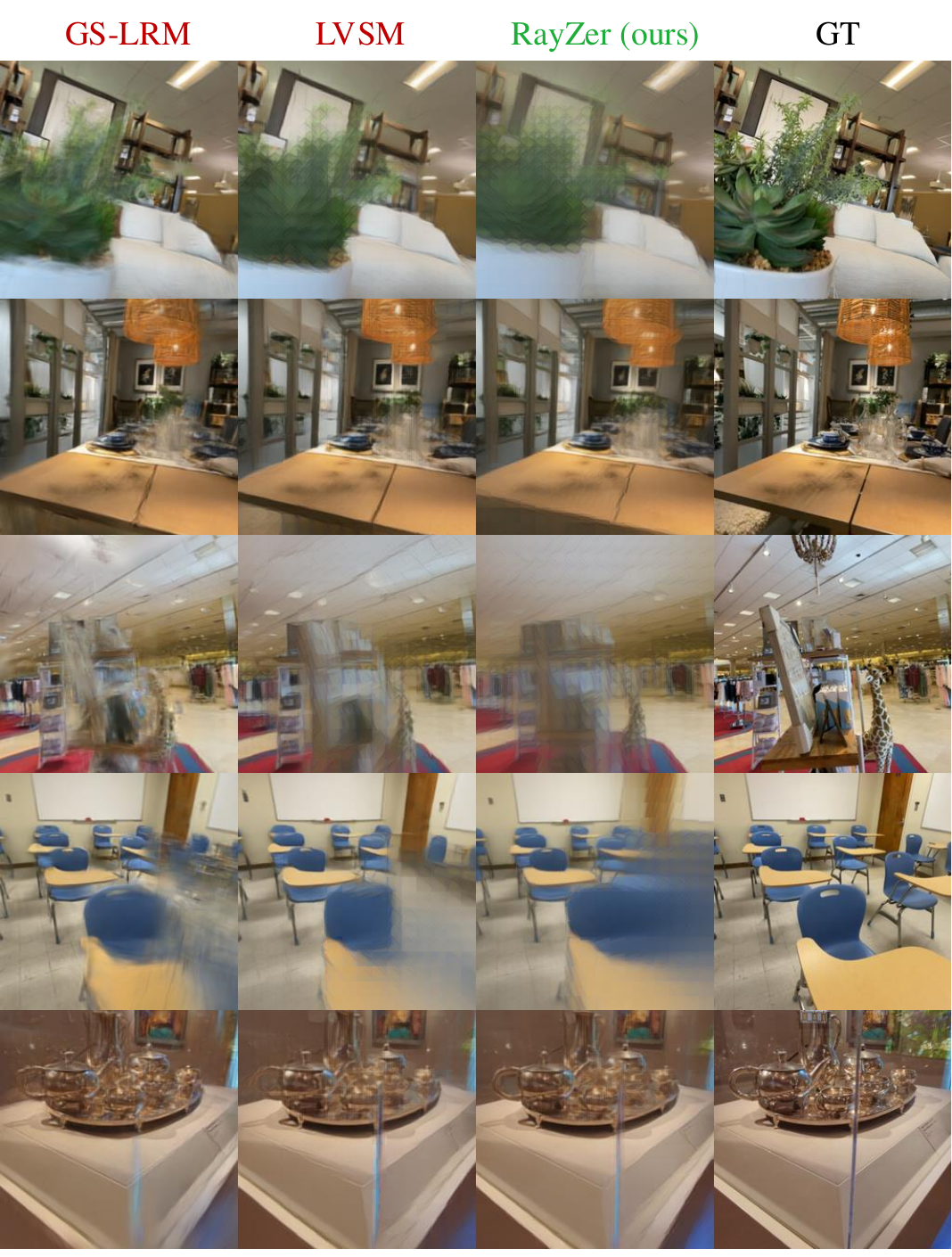}
    \vspace{-0.3in}
    \caption{\small{\textbf{Visualization of \modelname{} failure cases} on DL3DV.}} 
    \label{fig: failure}
    \vspace{-0.15in}
\end{figure}

\section{More Results}

In this section, we present more results for discussing \modelname{}'s failure cases and show more visualizations.

\noindent \textbf{Failure Case Pattern}. We observe that \modelname{} can fail when dealing with fine-grained geometry, complicated materials, and occlusions. We present the visualization in Fig.~\ref{fig: failure}. In detail, \modelname{} fails to handle complicated plant geometry (first row). This failure is not specific to \modelname{} -- GS-LRM and LVSM also can not handle it. In the second and last row, \modelname{} fails to handle multiple stacked glasses and is not perfect on the specular reflection of the silver teapots. GS-LRM and LVSM also demonstrate imperfect results. In the third and fourth rows, all methods, including \modelname{}, fail to handle occlusions, where the side view of the  exhibition stand is not observed in input views (third row), and the chairs in the fourth row have self-occlusion. 

\noindent \textbf{More Comparisons}. We present more visualization results, comparing with GS-LRM and LVSM in Fig.~\ref{fig: compare_supp}. \modelname{} generally performs on par, while being a self-supervised method that does not require any camera pose annotations.

\noindent \textbf{More Visualization}. We present more visualization results comparing with ground-truth novel views in Fig.~\ref{fig: compare_supp_gt}-\ref{fig: compare_supp_gt3}.

\section{More Discussion}
Why does \modelname{} demonstrates strong novel view synthesis quality while the fine-tuned pose estimation is not perfect (Table 7 in the main manuscript)? We conjecture \modelname{}'s pose space jointly learns the actual pose information and 3D-aware video frame interpolation at the same time. On datasets with small camera baselines (RealEstate), which is easy to learn, \modelname{} mainly focuses on learning actual pose estimation. This is supported by the accurate pose estimation performance on RealEstate. On datasets that have large camera baselines (DL3DV and Objaverse), where pose estimation is harder to learn with only self-supervision, \modelname{} also leverages video interpolation cues together with pose estimation to perform novel view synthesis. 

Thus, the method to further enhance disentanglement of interpolation and pose estimation would be an important future direction. In \modelname{}, using unordered image sets for training and using continuous video frames for training can be two extreme cases in the spectrum for learning this disentanglement. In detail, learning on continuous video frames with using image index positional embeddings strongly encourages the camera pose local smoothness to enhance training performance; while training on unordered image sets fully discards this prior. Finding a balance between the two and designing a better method to encourage the camera pose local smoothness is a promising avenue to solve the structure-and-motion problem with learning \texttt{SE(3)} camera poses in the real-world space.

\begin{figure}[t]
    \centering
    \includegraphics[width=\linewidth]{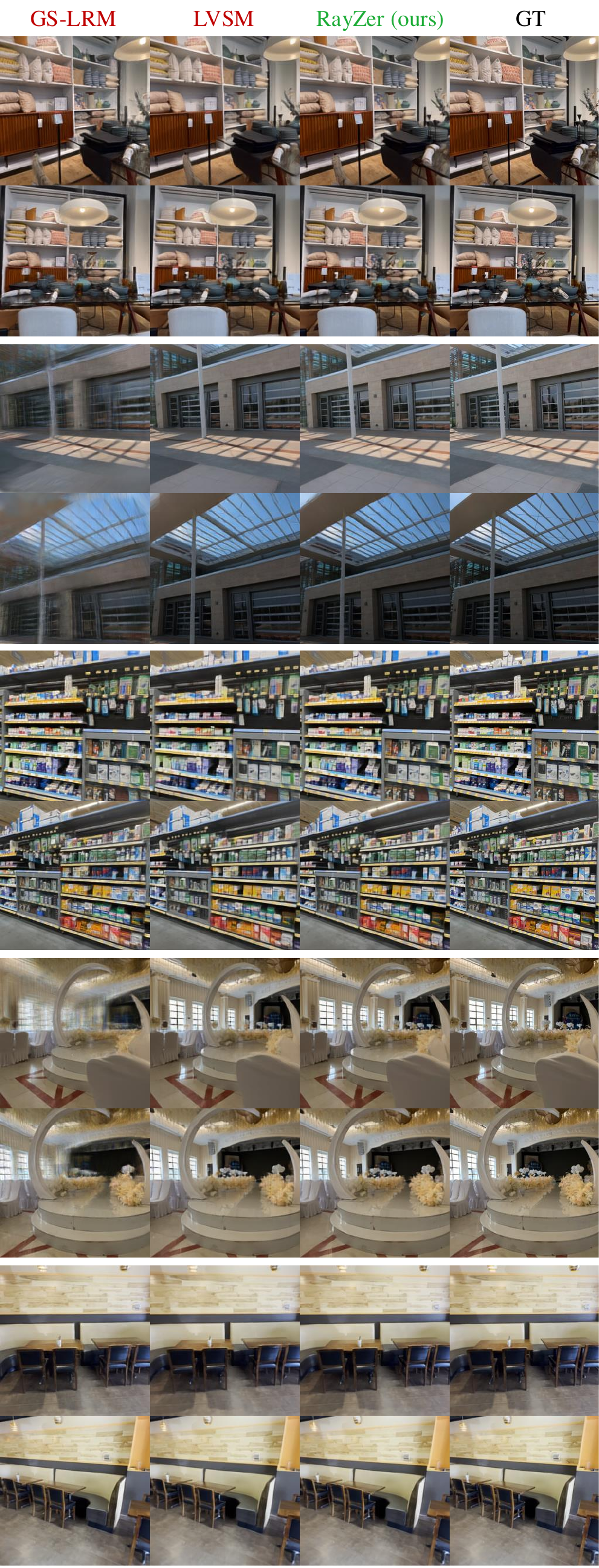}
    \vspace{-0.3in}
    \caption{\small{\textbf{Visual compression of \modelname{} and ``oracle'' methods} on DL3DV.}} 
    \label{fig: compare_supp}
    \vspace{-0.1in}
\end{figure}

\begin{figure*}[t]
    \centering
    \includegraphics[width=0.87\linewidth]{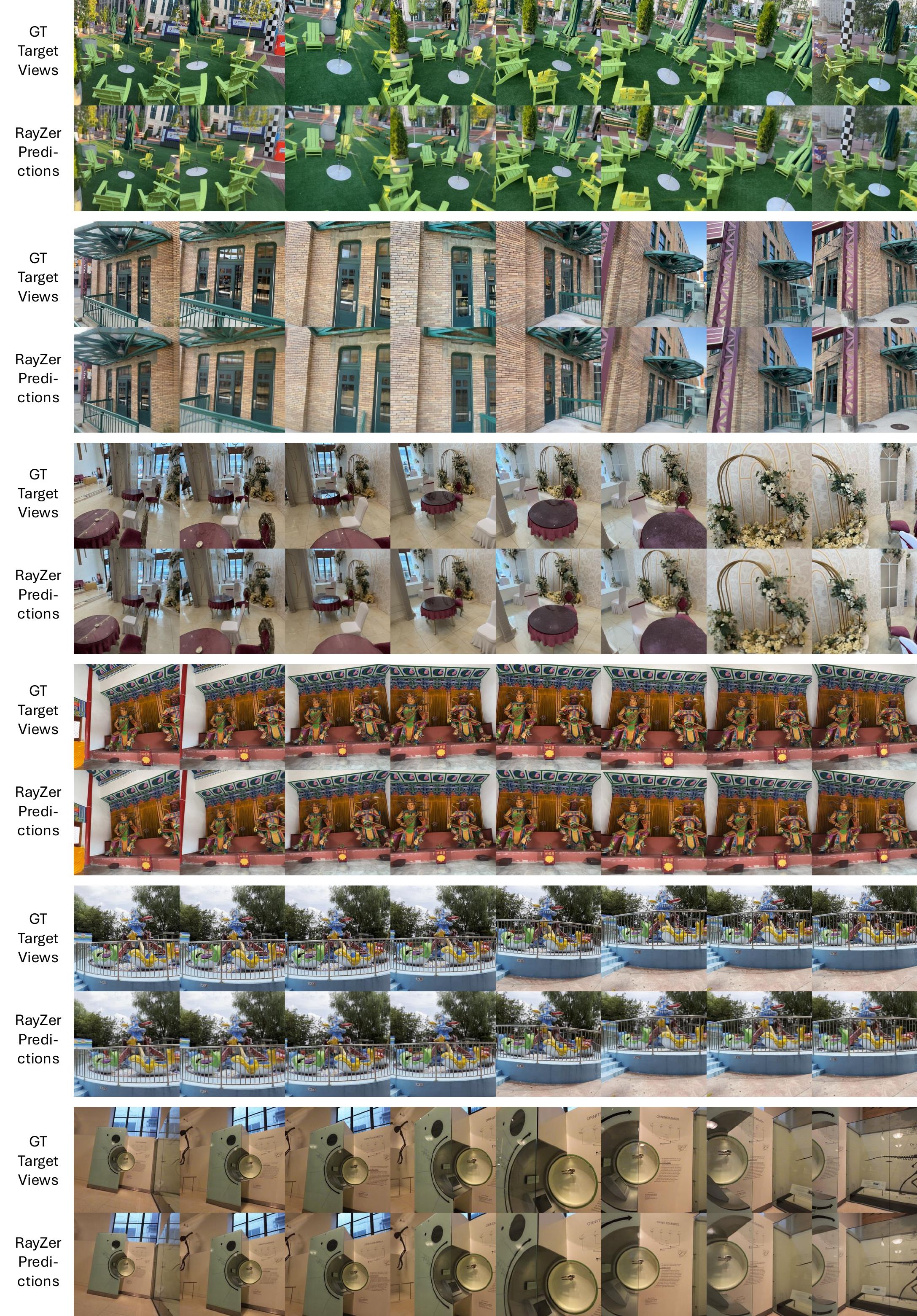}
    \vspace{-0.1in}
    \caption{\small{\textbf{Visual compression with ground-truth novel views} on DL3DV. The first row of each sample is the target novel views, and the second row are images rendered by \modelname{}.}} 
    \label{fig: compare_supp_gt}
    \vspace{-0.1in}
\end{figure*}

\begin{figure*}[t]
    \centering
    \includegraphics[width=0.93\linewidth]{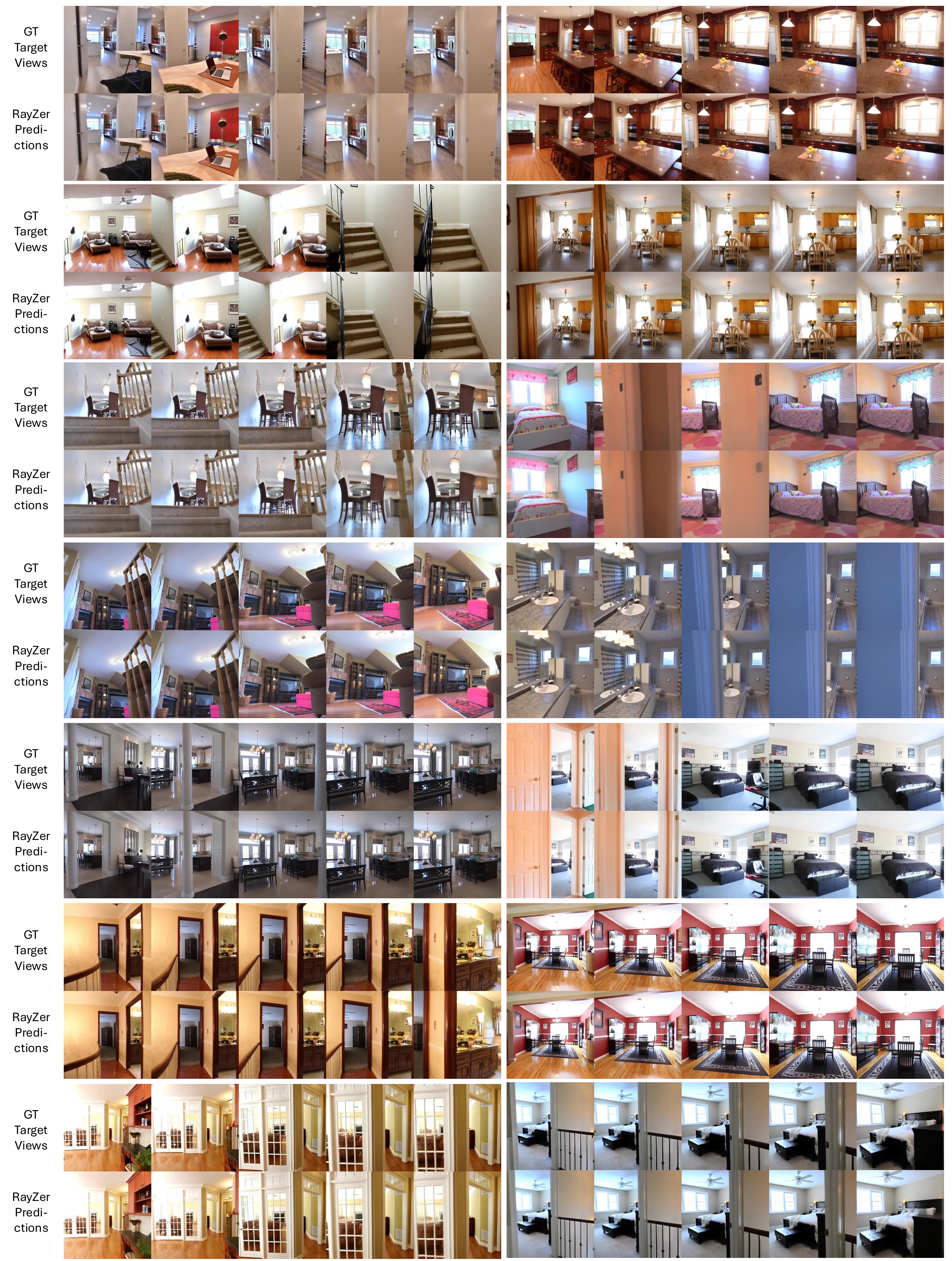}
    \vspace{-0.1in}
    \caption{\small{\textbf{Visual compression with ground-truth novel views} on RealEstate. The first row of each sample is the target novel views, and the second row are images rendered by \modelname{}.}} 
    \label{fig: compare_supp_gt2}
    \vspace{-0.1in}
\end{figure*}

\begin{figure*}[t]
    \centering
    \includegraphics[width=0.87\linewidth]{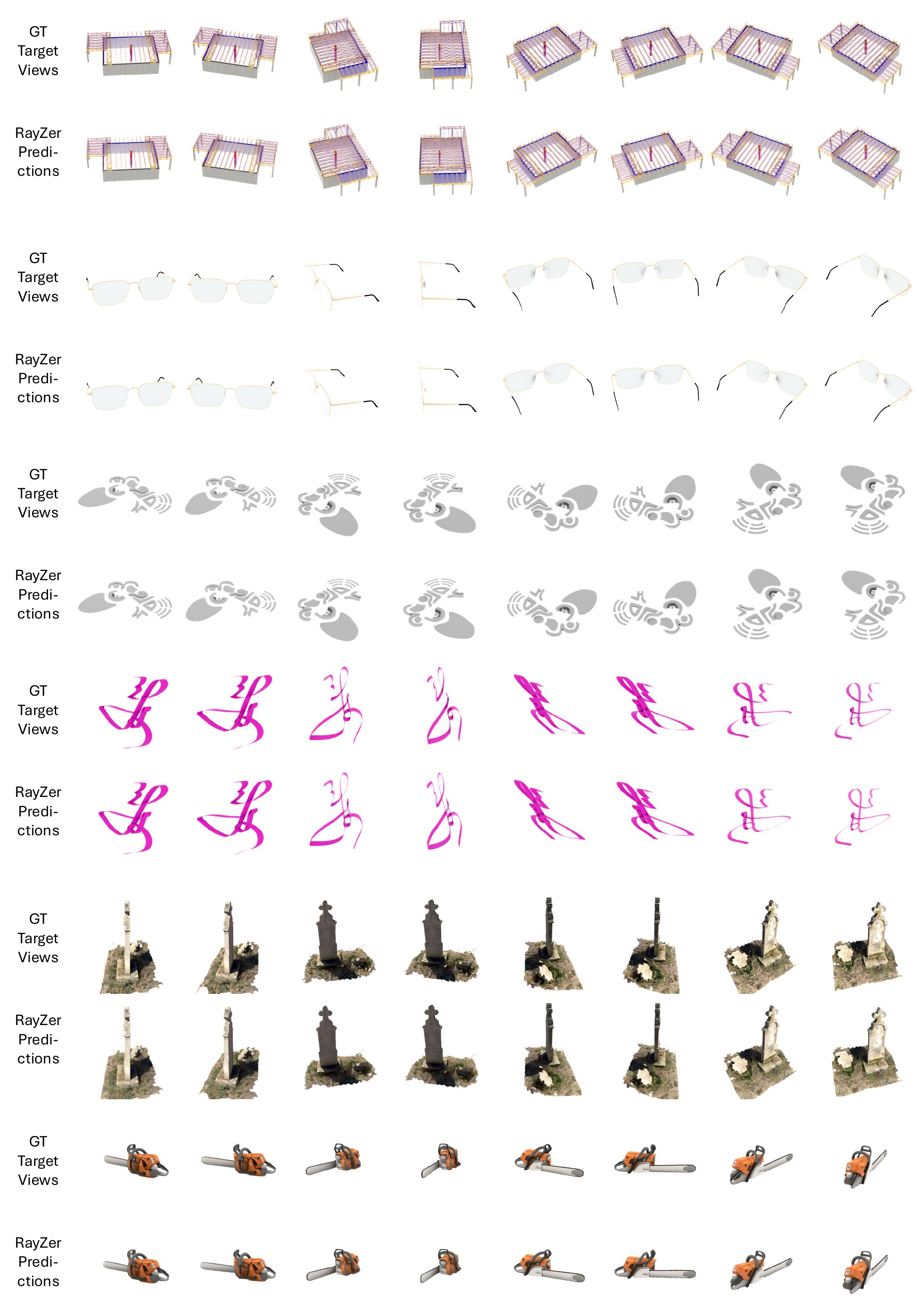}
    \vspace{-0.1in}
    \caption{\small{\textbf{Visual compression with ground-truth novel views} on Objaverse. The first row of each sample is the target novel views, and the second row are images rendered by \modelname{}.}} 
    \label{fig: compare_supp_gt3}
    \vspace{-0.1in}
\end{figure*}

\end{suppmat}

\end{document}